\documentclass[letterpaper, 10 pt, conference]{ieeeconf}  

\IEEEoverridecommandlockouts                              

\usepackage{graphicx} 

\title{\LARGE \bf
Picking a Conveyor Clean by an Autonomously Learning Robot
}

\author{Janne V. Kujala$^{1}$, Tuomas J. Lukka$^{1}$ and Harri Holopainen$^{1}$
\thanks{$^{1}$ZenRobotics Ltd, Vilhonkatu 5 A, FI-00100 Helsinki, Finland.
        {\tt\small firstname.lastname@zenrobotics.com}}%
}

\begin{document}
\maketitle
\thispagestyle{empty}
\pagestyle{empty}

\begin{abstract}

We present a research picking prototype
related to our company's industrial waste sorting application.
The goal of the prototype is to be as autonomous as possible
and it both calibrates itself and improves its picking with
minimal human intervention.

The system learns to pick objects better based on a feedback sensor
in its gripper and uses machine learning to
choosing the best proposal from a random sample produced by simple hard-coded
geometric models.

We show experimentally the system improving its picking autonomously by
measuring the pick success rate as function of time.

We also show how this system can pick a conveyor belt clean, depositing 70 out of
80 objects in a difficult to manipulate pile of novel objects into the correct
chute.

We discuss potential improvements and next steps in this direction.

\end{abstract}

\section{INTRODUCTION}

In this article, we describe our research prototype system that can
pick piled waste from a conveyor belt.  The motivation for
this prototype is grounded in the existing industrial robotic
application of our company: robotic waste sorting.

ZenRobotics' robots have been sorting waste on industrial waste
processing sites since 2014.  At one of our sites, 4200 tons of
construction and demolition waste has been processed. Of that waste,
2300 tons of metal, wood, stone and concrete objects have been
picked up from the conveyor by our sorting robots.  Performance of
the robot in this environment is critical for paying back the
investment. Currently the robots are able to identify, pick and
throw objects of up to 20 kg in less than 1.8 seconds, 24/7.
The current generation robot was taught to grasp objects using
human annotations and a
reinforcement learning algorithm as mentioned in \cite{lukkazenrobotics}.

Robotic recycling is rapidly growing, and is already transforming
the recycling industry. Robots' ability to recognize, grasp and
manipulate an extremely wide variety of objects is crucial. 
In order to provide this ability in a cost-effective way, new
training methods which do not rely on hardcoding or
human annotation will therefore be required.
For example, changing the shape of the gripper or adding degrees
of freedom might require all picking
logic to be rewritten or at least labor-intensive retraining unless the system
is able to learn to use the new gripper or degrees of freedom by itself.

We have chosen to tackle a small subproblem of the whole sorting
problem: learning to pick objects autonomously.
This problem differs from the more studied problems of "cleaning a table
by grasping" \cite{rao2010grasping} and bin picking
\cite{domae2014fast,holz2014active,nieuwenhuisen2013mobile}
in several ways: 1) The objects are
novel and there is a large selection of different objects. Objects
can be broken irregularly. In effect, anything can (and probably will)
appear on the conveyor eventually. 2) The objects are placed
on the conveyor belt by a random process and easily form random piles.
3) On the other hand, this problem is made slightly easier by the
fact that it is not necessary to be gentle to the objects; fragile objects will likely
have been broken by previous processes already. Scratching or colliding
with objects does not cause problems as long as the robot itself
can tolerate it (see Fig.~\ref{fig:gripper}).

Our solution starts with no knowledge of the objects and works completely autonomously
to learn how to make better pickups using feedback, for example from sensors in the gripper
like opening or force feedback.
In the following sections, we will first describe the system in detail, describe
our experiments with the system and conclude.

\section{DESCRIPTION OF THE SYSTEM}

In this section we describe our prototype system in detail.

\subsection{Hardware}

\begin{figure}[thpb]
  \centering
  \includegraphics[width=\columnwidth]{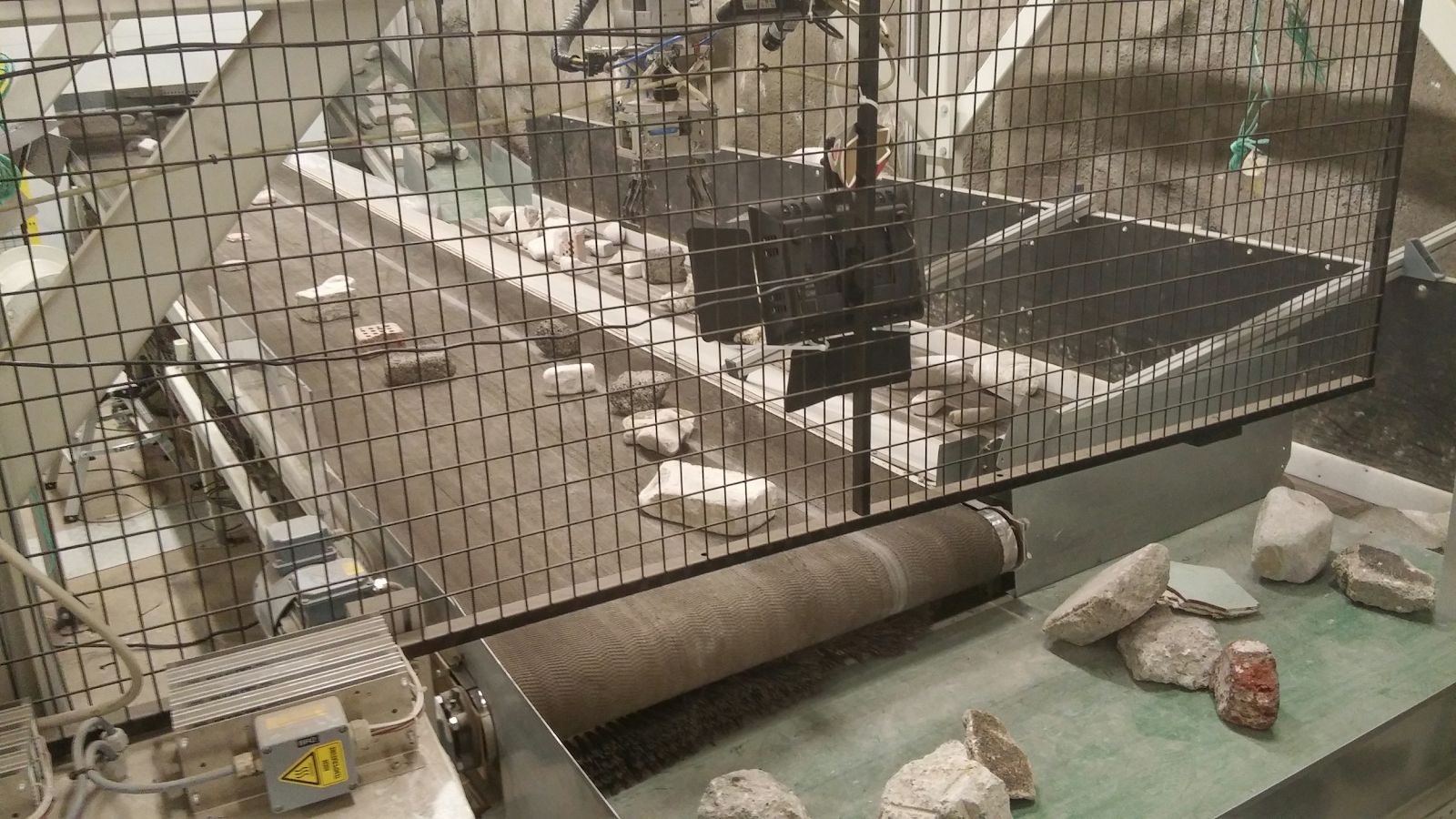}
  \caption{The waste merry-go-around used in the experiments to keep
    the material loop closed. The
    picked objects slide to the same conveyor as the other objects and
    all are brought back to the picking area with one more conveyor
    (occluded in this picture).}
  \label{fig:merry-go-around}
\end{figure}

\begin{figure}[thpb]
  \centering
  \includegraphics[trim = 520px 240px 610px 90px, clip=true, width=\columnwidth]{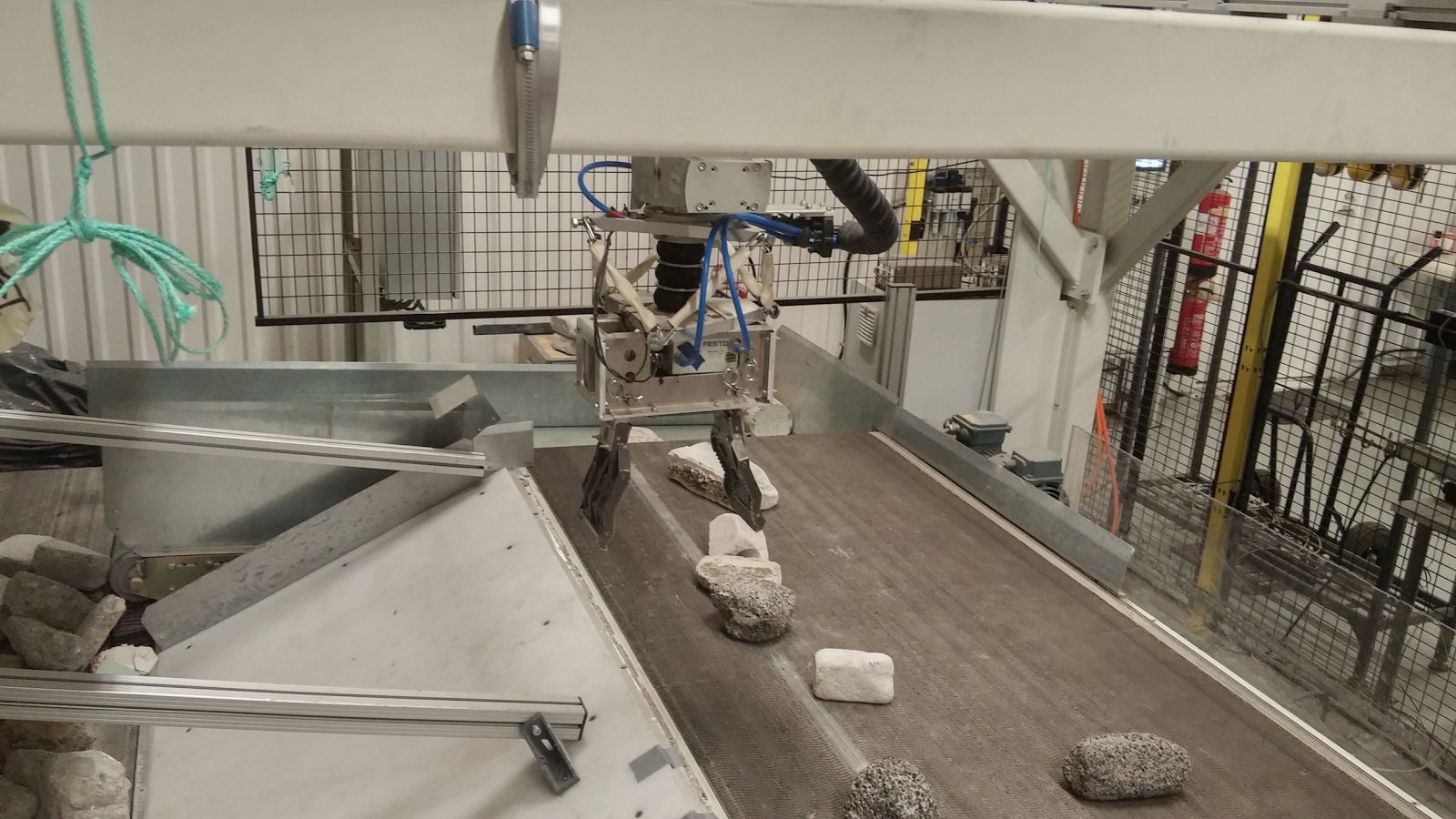}
  \caption{The gripper used in the experiments is an earlier version
    of our commercial gripper. This pneumatic gripper has a wide
    opening, it is position-controllable, and it contains a
    large-angle large-displacement compliance system while still being
    rigid when forces and torques do not exceed a threshold.}
  \label{fig:gripper}
\end{figure}

The hardware of our system consists of a waste merry-go-around
(Fig.~\ref{fig:merry-go-around}), a 3D camera (Asus Xtion), and a
gantry type robot (a prototype version of our product model).  The
gantry robot includes a wide-opening gripper and a large-angle
compliance system (Fig.~\ref{fig:gripper}).  The gripper has evolved
in previous versions of our product step by step to be morphologically
well-adapted to the task.

The gripper is position-controllable and has a sensor giving its
current opening.  In addition to the gripper opening, the robot has
four degrees of freedom, the $(x,y,z)$ coordinates and rotation around
the vertical axis (i.e., the gripper always faces down).

\subsection{SOFTWARE}

In our prototype system, we make use of our product's existing
software modules that handle conveyor tracking and motion planning to
execute a pick for a given \emph{handle}, a data structure similar to
the rectangle representation of Jiang et al. \cite{jiang2011efficient}
containing gripper $(x,y,z)$ coordinates, gripper angle, and gripper
opening for grasping an object.  In our prototype, we replace those
modules of our product that use information from line cameras to
decide where to grip.

\subsubsection{Automatic calibration}

\begin{figure}[thpb]
  \centering
  \includegraphics[width=\columnwidth]{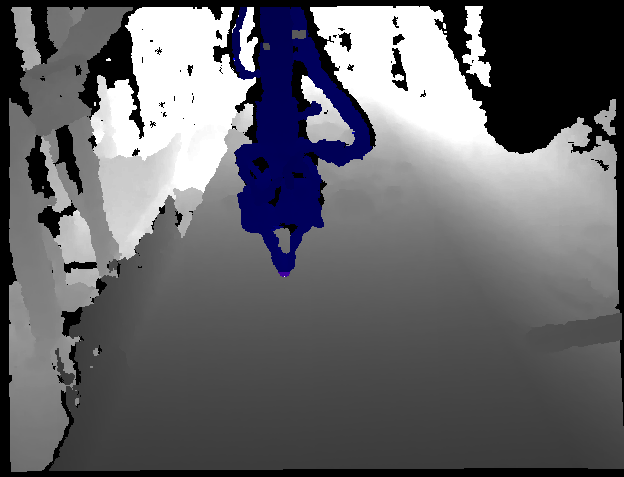}
  \caption{Automatic gantry coordinate system calibration: gripper is
    moved to a 3d grid of 60 different positions and the 3D camera
    $(x',y',z')$ coordinate of the tip of the closed gripper is detected
    from each position and stored with the corresponding gantry
    $(x,y,z)$ coordinates (the 3D camera image and detected gripper tip
    for one position is shown in the image). A projective
    transformation $(x', y', 1/z') \mapsto (x, y, z)$ is fitted to the
    data.}
  \label{fig:grippermask}
\end{figure}

Recently several methods have been developed (see
\cite{pradeep2014calibrating} and the references therein) for
calibrating sensors to robots.  For the present prototype, we use a
simplified automatic procedure for calibrating the 3D camera's
$(x',y',z')$ coordinates to the gantry $(x,y,z)$ coordinates
(Fig.~\ref{fig:grippermask}).  The gripper's angle and opening
parameters are calibrated separately using known gripper geometry parameters.
 
\subsubsection{Heightmap generation}

\begin{figure*}[thpb]
  \centering
  \includegraphics[width=\textwidth]{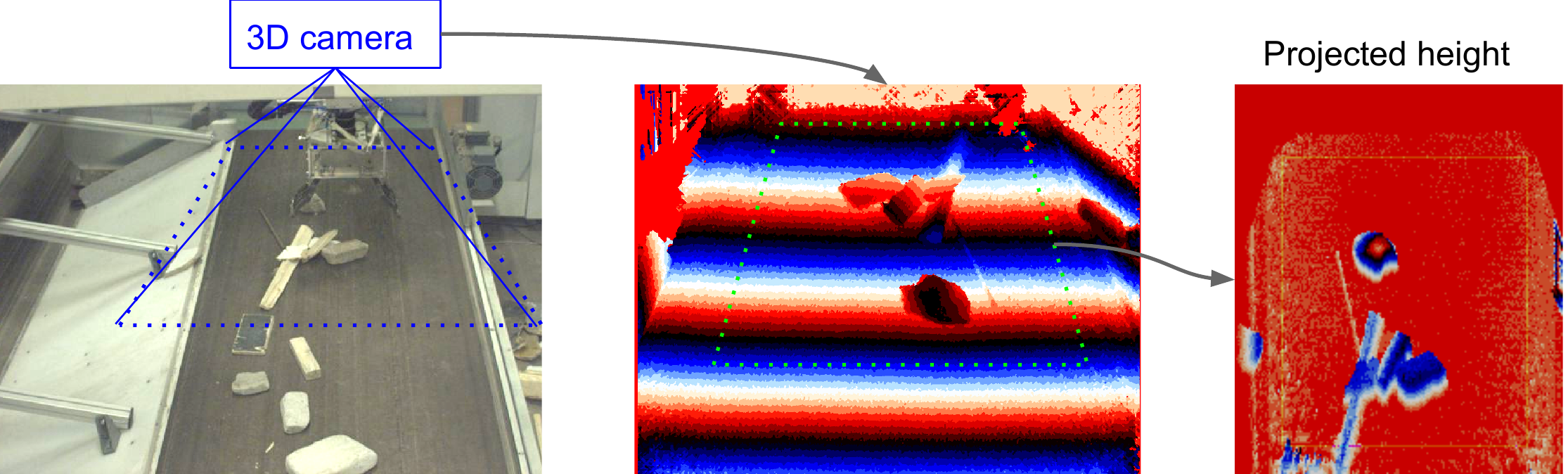}
  \caption{Heightmap generation: 3D camera image is projected into an
    isometric heightmap in gantry $(x,y)$ coordinates.}
  \label{fig:dcam-projection}
\end{figure*}

\begin{figure*}[thpb]
  \centering
  \includegraphics[width=\textwidth]{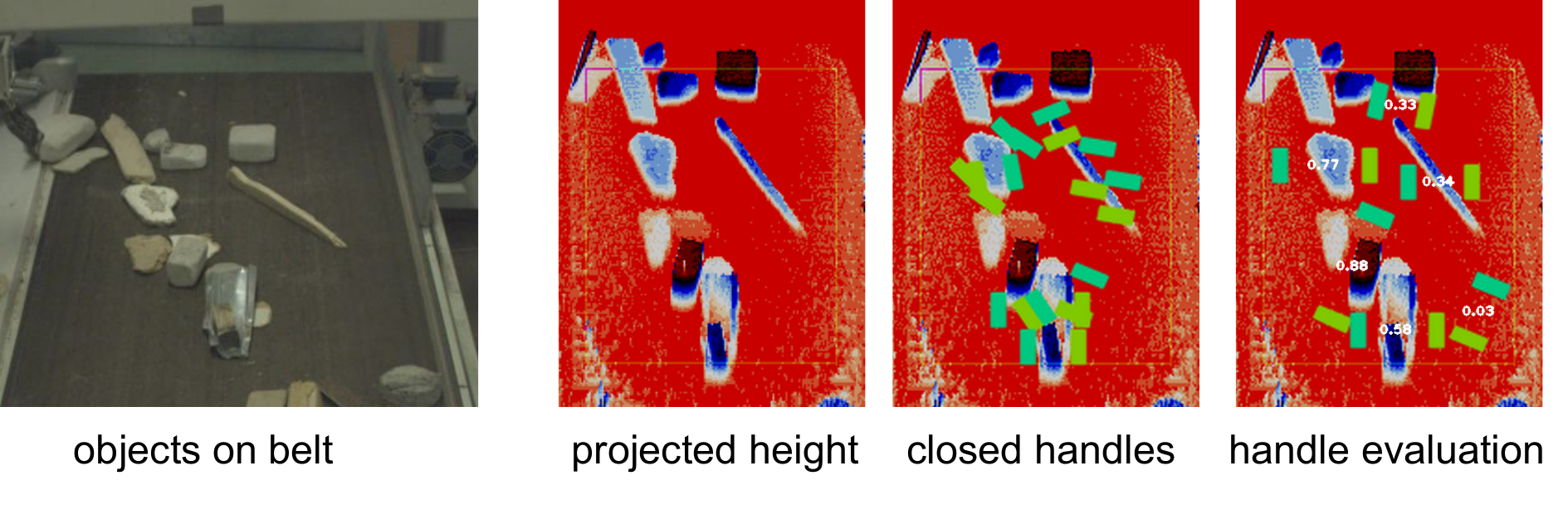}
  \caption{Handles are generated in two stages: first, all
    \emph{closed handles} are enumerated and a sample of size 200 is
    generated using probabilities proportional to the sum of the
    slopes at the finger contact points (a sample of size 10 is shown
    in the figure).  Then, based on a trained model and features
    calculated from the heightmap around the handle, each handle in
    the sample is evaluated for all possible extra openings.  The
    figure shows the estimated success probability (proportion of
    ``success'' votes from the random forest) for certain handles.
    The best handle is chosen for picking (except when its score is below 0.1 in which case it is only attempted with a 5\% probability).}
  \label{fig:handle-evaluation}
\end{figure*}

The 3D camera image\footnote{Figures \ref{fig:dcam-projection},
  \ref{fig:linesearch}, and \ref{fig:handle-evaluation} show depth
  images from an earlier version of our prototype using a higher
  resolution industrial Ensenso N20 depth sensor instead of the Asus Xtion
  that was used in the expreriments reported here.}
is projected using GPU into an isometric heightmap
defined on gantry $(x,y)$ coordinates
(Fig.~\ref{fig:dcam-projection}).  The projection code marks pixels
that are occluded by objects to their maximum possible heights and
additionally generates a mask indicating such unknown pixels.

\subsubsection{Handle generation}

\begin{figure*}[thpb]
  \centering
  \includegraphics[width=5in]{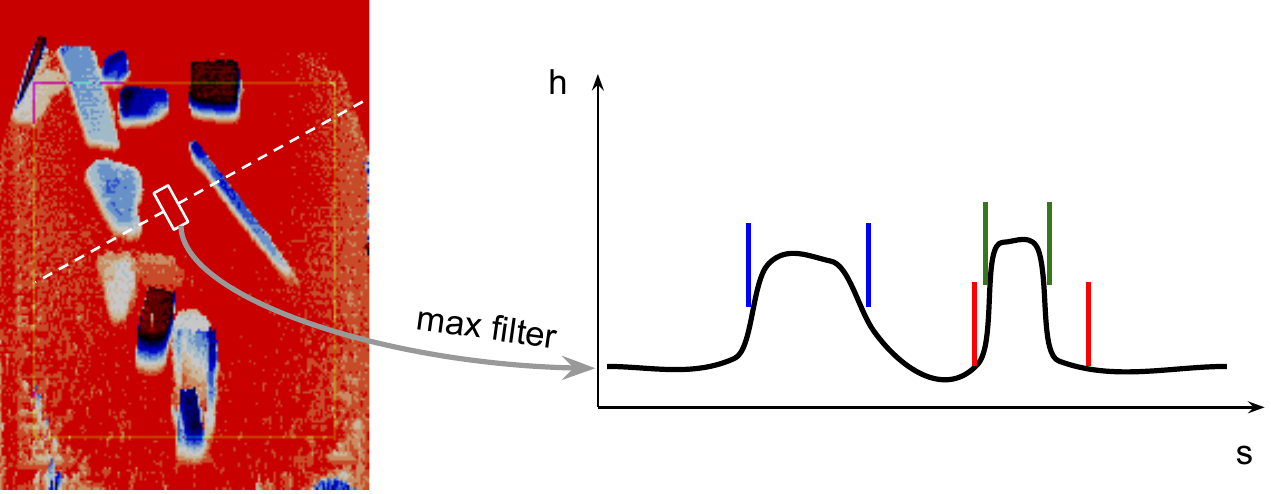}
  \caption{Exhaustive search of closed handles: a rectangular kernel
    of the shape of the gripper finger is moved across a line on the
    heightmap and yields (by maximum-filtering) a height curve $h(s)$
    indicating the minimum possible height of the gripper finger given
    the conveyor contents; closed handles aligned on the line are
    determined by pairs $(s_0, s_1)$ such that $h(s_0) < h(s) >
    h(s_1)$ for all $s_0 < s < s_1$ (three examples are shown in the
    figure); a stack-based algorithm for generating all closed handles
    over the line runs in linear time w.r.t. the number of pixels on
    the line.}
  \label{fig:linesearch}
\end{figure*}

The handle generation happens in two stages: first, we exhaustively
search through all \emph{closed handles}, that is, gripper
configurations where each finger of the gripper touches the heightmap
and the heightmap rises between the two points
(Fig.~\ref{fig:linesearch}).
The full set of closed handles are weighted by the sum
\[
  [h(s_0+1\textnormal{ pixel})-h(s_0)] + [h(s_1-1\textnormal{ pixel})-h(s_1)]
\] of height differences at the gripper contact points shown in
Fig.~\ref{fig:linesearch}.  A sample of 200 handles is generated
using probabilities proportional to the weights.  After this, each
handle in the sample is duplicated for all possible extra-openings
allowed by the heightmap (taking into account the nonlinear movement
of the gripper as it opens and closes) and the maximum opening of the
gripper.  This completes the hard-coded stage of handle generation.

For every handle of the first stage, features are generated from the
heightmap around the handle.  The features are based on
\begin{itemize}
  \item $80\times39$ pixel ($40\times19.5$ cm) slices of the heightmap aligned at the left
    finger, center, and right finger of the gripper (including a margin of 4 cm around the rectangle
    inside the gripper fingers),
  \item the opening of the handle and extra opening to be applied when grasping, and
  \item the height of the handle (which is subtracted from the
    heightmap slices so as to yield translation invariant features).
\end{itemize}
Of these, the image features are further downsampled and transformed
by a dual-tree complex wavelet transform \cite{selesnick2005dual} to yield the
inputs for a random forest that is trained to classify
the handles into those that succeed and those that fail.  The handle that
gets the best score (most votes from the random forest) is chosen for
picking (except when its score is below 0.1 in which case it is only attempted with a 5\% probability in order to avoid picking the empty belt for aesthetic reasons).  When there is no trained model available, a random handle
from the output of the first stage is chosen for picking.

\subsubsection{Feedback for autonomous training}

During each picking attempt, the system monitors the gripper opening
and if the gripper closes (almost) completely before completing the
throw, it is determined that the object has slipped and the pick is
aborted.  This post-verification signal yields the necessary feedback
for training.

The features and result of each pick attempt are stored and a
background process reads these training samples periodically and
trains a new handle model based on all collected data.  When a new model is trained, the system
starts using it on the next pick attempt.

The immediate feedback from failed and successful attempts allows the
system to learn quickly and autonomously and to adapt to novel
objects.

\section{EXPERIMENTS}

\subsection{Autonomously learning to pick}

In this experiment, the conveyor under the system was cleared for
calibration, the calibration was run, and the conveyor was started at a
slow constant speed.  When there were objects coming under the robot,
the picking software was started.  The system started picking with
just the hard-coded first stage model.  After every 100 pick attempts, the system
trained the second-stage model using data from all pick attempts from
the beginning and started using the newly trained model on subsequent
picks.  For technical reasons related to data collection,
the system was paused briefly every 15 minutes.

\begin{figure}[thpb]
  \centering
  a)\parbox{2.7in}{\includegraphics[width=2.7in]{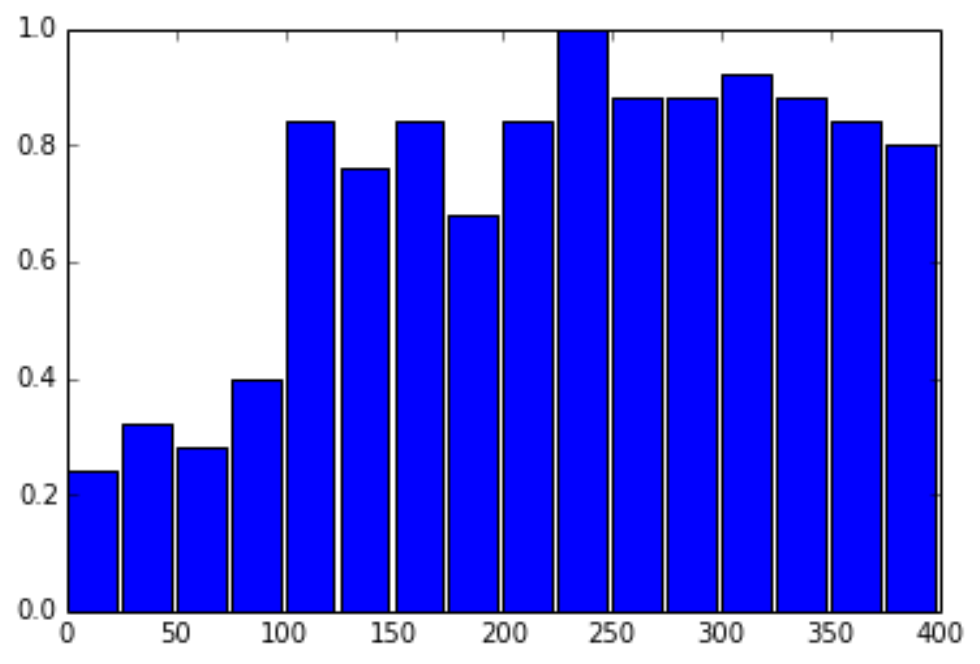}}

  \vspace{5pt}
  b)\parbox{2.7in}{\includegraphics[width=2.7in]{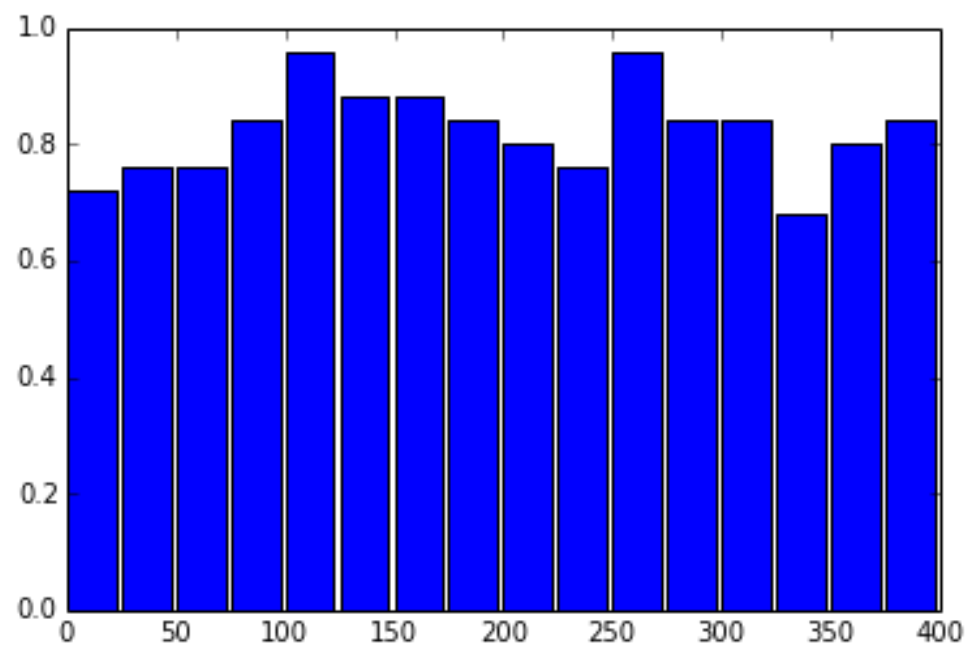}}
  \caption{Autonomous learning: proportion of successful picks in
    blocks of 25 pick attempts when starting from no knowledge and
    training a) every 100 pick attempts or b) every 10 seconds.  It is
    evident from a) that there is a clear improvement of the success
    rate after the model is trained and from b) that using immediate
    feedback, the learning can be very fast.  }
  \label{fig:exper23}
\end{figure}

The results of this experiment are shown in Fig.~\ref{fig:exper23}a.
The same experiment was repeated running the training every 10
seconds.  The results are shown in Fig.~\ref{fig:exper23}b.  From
these results, it is clear that the immediate feedback from
post-verification allows autonomous learning that can be very fast.

\subsection{Picking the conveyor clean}

\begin{figure}[thpb]
  \centering
  \includegraphics[width=\columnwidth]{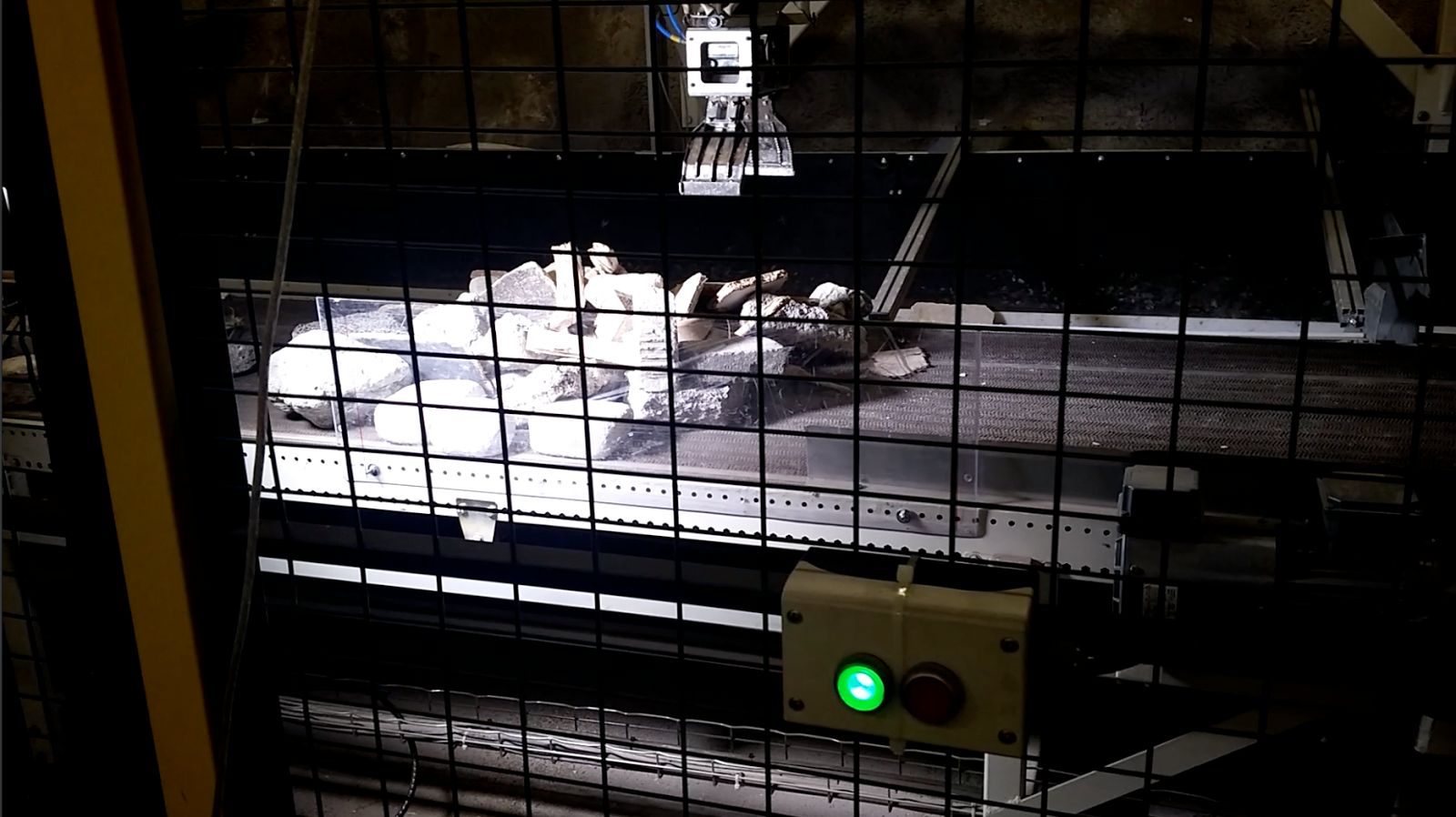}

  \vspace{5pt}
  \includegraphics[width=\columnwidth]{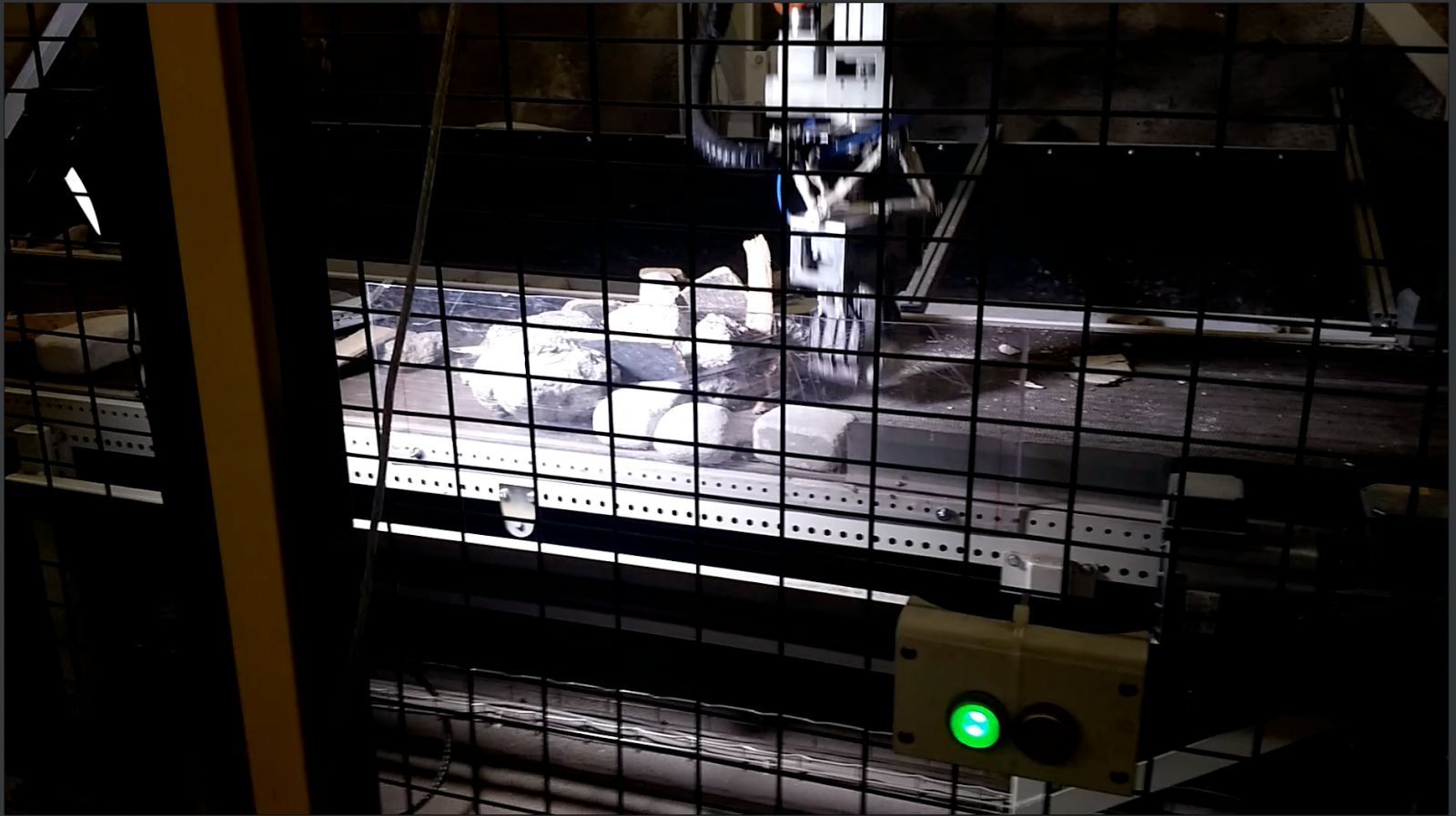}

  \vspace{5pt}
  \includegraphics[width=\columnwidth]{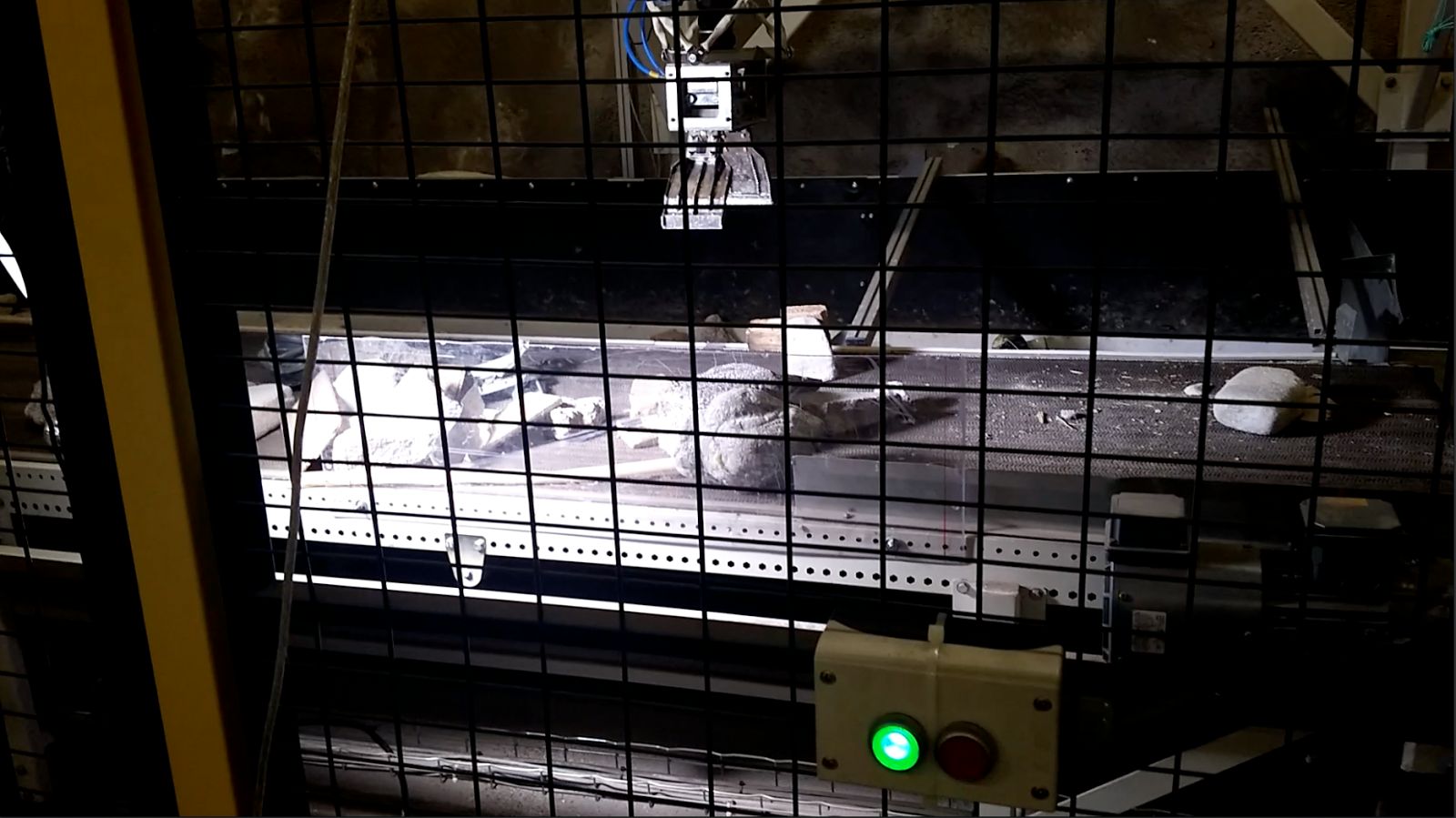}

  \vspace{5pt}
  \includegraphics[width=\columnwidth]{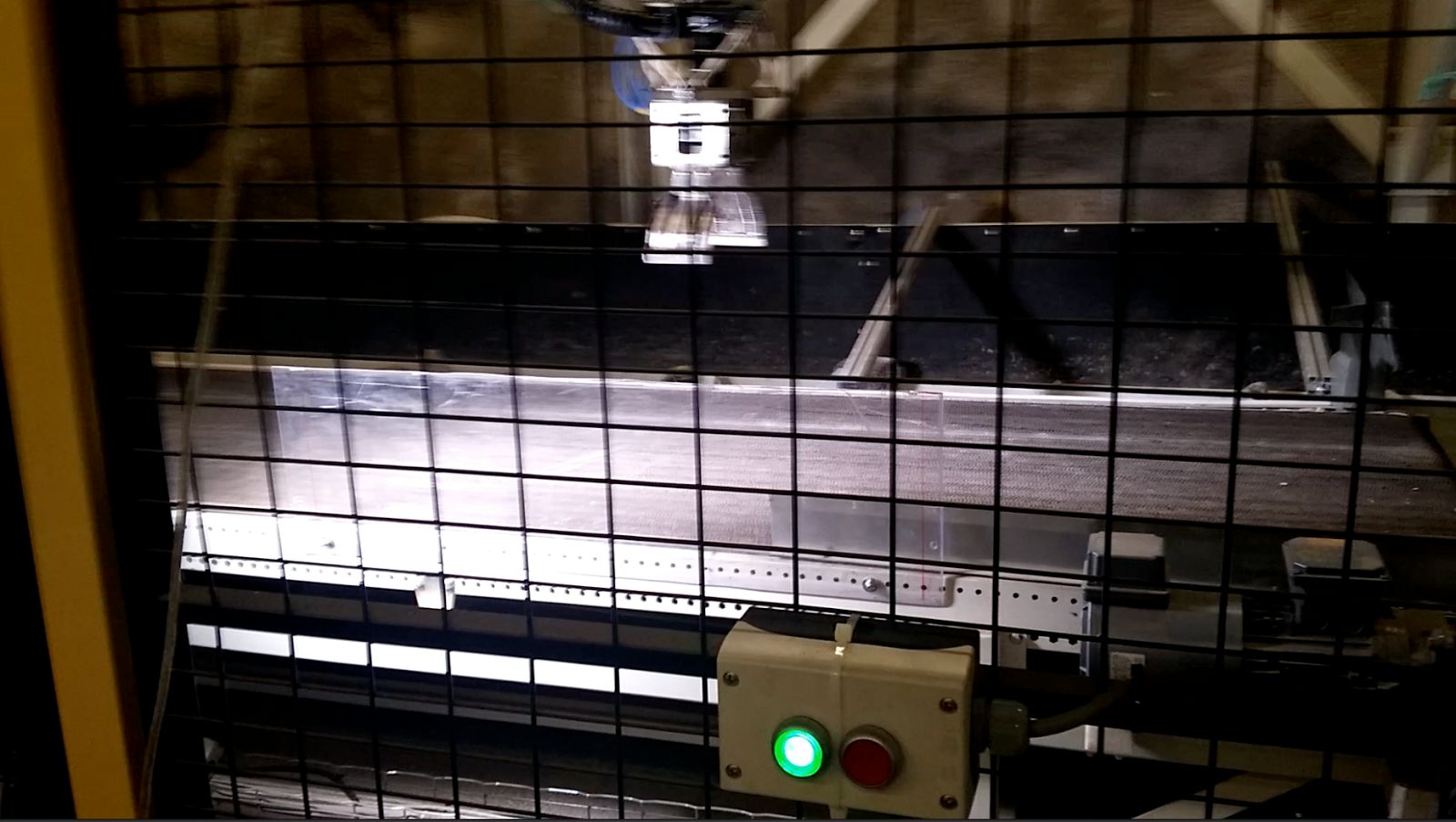}
  \caption{Picking the conveyor clean. Some shots from the accompanying
    video, after an initial learning period.
    By our count from the video of the experiment, 70 out of
    80 objects were correctly deposited in the right chute.
    The third frame shows on the right one of the objects that slipped beyond the
    working area by failed pickup attempts.  }
  \label{fig:emptying}
\end{figure}

In this experiment, the conveyor under the system was cleared for
calibration, the calibration was run, and after moving the conveyor until
there were objects in the working area, the picking software was
started.  Then, the conveyor movement was controlled manually, moving it
short distances at a time, so as to let the robot pick the conveyor clean.
The system started picking using just the hard-coded first stage model
and the second stage model was trained on data from all picking
attempts from the beginning every 10 seconds.  The picking performance
improved during the experiment as in the other experiments.  Although
somewhat more pick attempts will fail than on a constantly moving
conveyor, the system will retry picking any objects left on the working
area until it succeeds.  The accompanying video shows how, after some
training, the system clears a large pile from the conveyor
(Fig.~\ref{fig:emptying}).

\section{CONCLUSION AND FUTURE WORK}

We have demonstrated a prototype system that is able to pick a pile of novel waste objects from a conveyor
and which has autonomously learned to select better points to pick from. We have shown that performing this task with a 4-dof
robot with a single camera not on top of the system is possible.

It is easy to think of several ways to improve the performance of the system. For the picking the conveyor clean -task,
simply adding better edges to the conveyor and making the working area slightly larger would help - currently the working
area is very limited due to the 3D camera used. The machine learning algorithm used is very simple. Enlarging the set of candidate handles
could boost performance significantly and would be easy to parallelize on the GPU.
It would also be possible to make the hard-coded first stage
less conservative regarding shadows.

On the other hand, it would be possible to address some of the specific types of errors that were observed:
\begin{itemize}
  \item grasping shadow: our current handle model does not make use of
    the mask indicating areas with unknown height (i.e., areas
    occluded by objects from the 3D camera's point of view); using
    this information in the features would allow learning to better
    handle the shadows; alternatively two 3D cameras could be used to
    reduce shadows
  \item grasping at object (corner) that just came in range: this
    could be improved by additional logic to avoid handles at the edge
  \item grasping at empty belt: when there are no objects, small
    variations of the conveyor height, small particles, or sensor noise may yield
    handles; we have reduced such pick
    attempts by avoiding picking (except by small probability) when the score of the best
    handle is below certain threshold
  \item thin objects: the postverification may yield incorrect failure
    signal when grasping a thin object and the system may learn to
    avoid picking thin objects; this shows the importance of the
    feedback signal
  \item heavy stones slipping: could use slower throw, adding throw
    acceleration as another degree of freedom for the generated
    handles.
\end{itemize}

On the other hand, with this system, the point of diminishing returns is quickly reached because the system can retry picks
that failed. The difference between an 80\% success rate and 90\% success rate is relatively minor, as opposed to 
the same difference in a line scanning system where 80\% would mean double the number of unpicked objects from 90\%.

At the moment, the cycle time of the prototype, around 6 seconds, is a far cry from our production system's
1.8 s cycle time. However, there is no fundamental reason why such a cycle time could not be reached by this type of system;
the difference is mostly caused by the prototype being very conservative about when the images are taken and not being yet optimized.

More interesting extensions of the systems in terms of practical applicability would be, e.g, learning to control the conveyor
in order to maximize some function of the amount of picked material and the percentage of objects that get picked; sorting
objects by some characteristic while picking, and learning to carefully pick one object at a time. In the current setup,
the last one was not a problem; two-or-more-object picks were rare but this may be more related to the size of the objects and the gripper.

\section{ACKNOWLEDGMENTS}

The authors would like to thank the ZenRobotics team of research assistants for helping in this work, especially Risto Sirvi{\"o} for supervising many of the experiments and Risto Sirvi{\"o} and Sara Vogt for annotating experiment data.  The authors would also like to thank Risto Bruun, Antti Lappalainen, Arto Liuha, and Ronald Tammep{\~o}ld for discussions and PLC work, Timo Tossavainen for many discussions, and Risto Bruun, Juha Koivisto, and Jari Siitari for hardware work.  This work also makes use of the contributions of the whole ZenRobotics team through the parts of our product that were reused in this prototype.

\bibliographystyle{IEEEtran}

\end{document}